\documentclass[5p]{elsarticle}

\usepackage[frozencache,cachedir=.]{minted}
\usepackage{xcolor}
\usepackage{hyperref}

\definecolor{lightgray}{rgb}{0.95, 0.95, 0.95}
\definecolor{console}{rgb}{0.4, 0.4, 0.4}

\usepackage{amssymb}

\begin{document}

\begin{frontmatter}



\title{PySS3: A Python package implementing a novel text classifier \\with visualization tools for Explainable AI}

\author[unsl,conicet]{Sergio G. Burdisso\corref{cor}}
\ead{sburdisso@unsl.edu.ar}

\author[unsl]{Marcelo Errecalde}
\ead{merreca@unsl.edu.ar}

\author[inaoe]{Manuel Montes-y-G\'omez}
\ead{mmontesg@inaoep.mx}

\address[unsl]{Universidad Nacional de San Luis (UNSL), Ej\'ercito de Los Andes 950, San Luis, San Luis, C.P. 5700, Argentina}
\address[conicet]{Consejo Nacional de Investigaciones Cient\'ificas y T\'ecnicas (CONICET), Argentina}
\address[inaoe]{Instituto Nacional de Astrof\'isica, \'Optica y Electr\'onica (INAOE), Luis Enrique Erro No. 1, Sta. Ma. Tonantzintla, Puebla, C.P. 72840, Mexico}

\begin{abstract}
A recently introduced text classifier, called SS3, has obtained state-of-the-art performance on the CLEF's eRisk tasks. SS3 was created to deal with risk detection over text streams and, therefore, not only supports incremental training and classification but also can visually explain its rationale. However, little attention has been paid to the potential use of SS3 as a general classifier. We believe this could be due to the unavailability of an open-source implementation of SS3. In this work, we introduce PySS3, a package that implements SS3 and also comes with visualization tools that allow researchers to deploy robust, explainable, and trusty machine learning models for text classification.
\end{abstract}

\begin{keyword}
Text classification \sep XAI \sep SS3



\end{keyword}

\end{frontmatter}



\section{Introduction}
\label{sec:introduction}


A challenging scenario in the machine learning field is the one referred to as ``early classification''. Early classification deals with the problem of classifying data streams as early as possible without having a significant loss in performance.
The reasons behind this requirement of ``earliness'' could be diverse, but the most important and interesting case is when the classification delay has negative or risky implications.
This scenario, known as ``early risk detection'' (ERD), has gained increasing interest in recent years with potential applications in rumor detection \citep{ma2015detect,ma2016detecting,kwon2017rumor}, sexual predator detection, aggressive text identification \citep{escalante2017early}, depression detection \citep{losada2017erisk, losada2016test}, and terrorism detection \citep{iskandar2017terrorism}, among others.

A recently introduced machine learning model for text classification~\citep{burdisso2019}, called SS3, has shown to be well suited to deal with ERD problems on social media streams.
It obtained state-of-the-art performance on early depression, anorexia and self-harm detection on the CLEF eRisk open tasks~\citep{burdisso2019, burdisso2019-tss3, burdisso2019clef}.
Unlike standard classifiers, this new classification model was specially designed to deal with ERD problems since: it can visually explain its rationale, and it naturally supports incremental training and classification over text streams.
Moreover, SS3 introduces a classification model that does not require feature engineering and is robust to the Class Imbalance Problem, which has become one of the most challenging research problems~\cite{zhang2019multi}.

However, little attention has been paid to the potential use of SS3 as a general classifier for document classification tasks.
One of the main reasons could be the fact that there is no open-source implementation of SS3 available yet.
We believe that the availability of open-source implementations is of critical importance to foster the use of new tools, methods, and algorithms.
On the other hand, Python is a popular programming language in the machine learning community thanks to its simple syntax and a rich ecosystem of efficient open-source implementations of popular algorithms.

In this work, we introduce ``PySS3'' and share it with the community. PySS3 is an open-source Python package that implements SS3 and comes with two useful tools that allow working with it in an effortless, interactive, and visual way. For instance, one of these tools provides post-hoc explanations using visualizations that directly highlight relevant portions of the raw input document, allowing researchers to better understand the models being deployed by them. Thus, PySS3 allows researchers and practitioners to deploy more robust, explainable, and trusty machine learning models for text classification.

\section{Background}
\label{sec:background}

In this section, we provide an overview of the SS3 classifier. We will introduce only the general idea and basic terminology needed to understand the PySS3 package better. Readers interested in the formal definition of the model are invited to read Section 3 of the original paper\cite{burdisso2019}.

\subsection{The SS3 classification model}

As it is described in more detail by Burdisso et al. \cite{burdisso2019}, during training and for each given category, SS3 builds a dictionary to store word frequencies using all training documents of the category.
This simple training method allows SS3 to support online learning since when new training documents are added, SS3 simply needs to update the dictionaries using only the content of these new documents, making the training incremental.
Then, using the word frequencies stored in the dictionaries, SS3 computes a value for each word using a function, $gv(w,c)$, to value words in relation to categories. 
$gv$ takes a word $w$ and a category $c$ and outputs a number in the interval [0,1] representing the degree of confidence with which $w$ is believed to \emph{exclusively} belong to $c$, for instance, suppose categories $C= \{food, music, health, sports\}$, we could have:

\vspace{5mm}
\begin{tabular}{l l} 
$gv(apple, tech) = 0.8;$ & $gv(the, tech) = 0;$\\
$gv(apple, business) = 0.4;$ & $gv(the, business) = 0;$\\
$gv(apple, food) = 0.75;$ & $gv(the, food) = 0;$
\end{tabular}
\vspace{5mm}

Additionally, a vectorial version of $gv$ is defined as: $$\overrightarrow{gv}(w)=(gv(w,c_0), gv(w,c_1), \dots, gv(w,c_k))$$ where $c_i \in C$ (the set of all the categories).
That is, $\overrightarrow{gv}$ is only applied to a word and it outputs a vector in which each component is the word's \emph{gv} for each category $c_i$.
For instance, following the above example, we have:

\vspace{5mm}
$\overrightarrow{gv}(apple) = (0.8, 0.4, 0.75)$

$\overrightarrow{gv}(the) = (0, 0, 0)$
\vspace{5mm}

The vector $\overrightarrow{gv}(w)$ is called the ``\emph{confidence vector} of $w$''.
Note that each category is assigned to a fixed position in $\overrightarrow{gv}$. For instance, in the example above $(0.8, 0.4, 0.75)$ is the \emph{confidence vector} of the word ``apple'' and the first position corresponds to $technology$, the second to $business$, and so on.

\subsubsection{Classification Process}
\label{subsec:classification}

The classification algorithm can be thought of as a 2-phase process.
In the first phase, the input is split into multiple blocks (e.g., paragraphs), then each block is repeatedly divided into smaller units (e.g., sentences, words). Thus, the previously ``flat'' document is transformed into a hierarchy of blocks.
In the second phase, the $\overrightarrow{gv}$ function is applied to each word to obtain a set of word \emph{confidence vectors}, which are then reduced to sentence \emph{confidence vectors} by a word-level \emph{summary operator}.
This reduction process is recursively propagated to higher-level blocks, using higher-level \emph{summary operators},\footnote{By default, in PySS3, the \emph{summary operators} are vector additions. However, PySS3 provides an easy way for the user to define his/her custom \emph{summary operators}. More info on this can be found here: \href{https://pyss3.rtfd.io/en/latest/user_guide/ss3-classifier.html}{\url{https://pyss3.rtfd.io/en/latest/user_guide/ss3-classifier.html}}} until a single \emph{confidence vector}, $\overrightarrow{d}$, is generated for the whole input.
Finally, the actual classification is performed by applying some policy based on the \emph{confidence values} stored in $\overrightarrow{d}$ ---for instance, selecting the category with the highest \emph{confidence value} in $\overrightarrow{d}$.

It is worth mentioning that it is quite straightforward to visually explain the classification process if different input blocks are colored proportionally to their \emph{confidence} values. As described in \autoref{sec:pyss3}, this characteristic is exploited by the ``Live Test'' tool to create interactive visual explanations.

\subsubsection{The Hyperparameters}
\label{subsec:ss3-hyperparameters}

The entire classification process depends on the $gv$ function since it is used to create the first set of \emph{confidence vectors} upon which higher-level \emph{confidence vectors} are then created.
As described in more detail in the original paper\cite{burdisso2019}, the computation of $gv$ involves three functions, $lv$, $sg$ and $sn$, as follows:

$$gv(w, c) = lv_\sigma(w, c)\cdot sg_{\lambda}(w, c)\cdot sn_\rho(w, c)$$

\begin{itemize}
\item $lv_\sigma(w, c)$ values a word based on the local frequency of $w$ in $c$. As part of this process, the word distribution curve is smoothed by a factor controlled by the hyperparameter $\sigma$.
\item $sg_{\lambda}(w, c)$ captures the significance of $w$ in $c$. It is a sigmoid function that returns a value close to $1$ when $lv(w, c)$ is significantly greater than $lv(w, c_i)$, for most of the other categories $c_i$; and a value close to $0$ when $lv(w, c_i)$ values are close to each other, for all $c_i$.
The $\lambda$ hyperparameter controls how far $lv(w, c)$ must deviate from the median to be considered significant.
\item $sn_\rho(w, c)$ decreases the global value in relation to the number of categories $w$ is significant to. That is, the more categories $c_i$ to which $sg_{\lambda}(w, c_i)\approx 1$, the smaller the $sn_\rho(w, c)$ value. The $\rho$ hyperparameter controls how severe this sanction is.
\end{itemize}

\section{PySS3}
\label{sec:pyss3}

\subsection{Software architecture}

\begin{figure*}[t!]
    \centering
    \includegraphics[width=190mm]{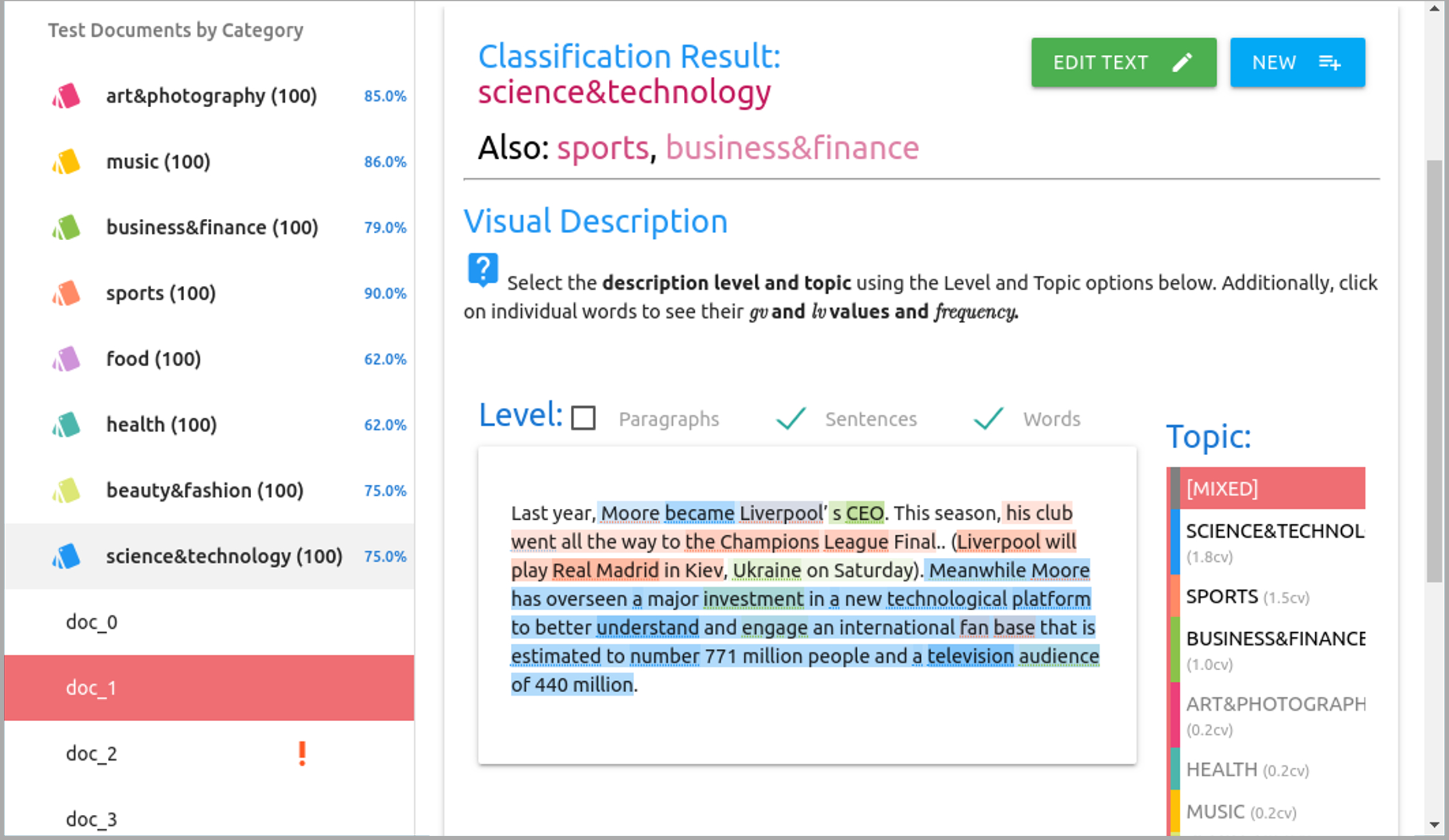}
    \caption{Live Test screenshot. On the left side, the list of test documents grouped by category is shown along with the percentage of success.
    Note the ``doc\_2'' document is marked with an exclamation mark (!); this mark indicates it was misclassified, which eases error analysis.
    The user has selected the ``doc\_1'', the ``classification result'' is shown above the visual description. 
    In this figure, the user has chosen to display the visual explanation at sentence-and-word level, using mixed topics.
    For instance, the user can confirm that, apparently, the model has learned to recognize important words and that it has correctly classified the document.
    Also, by using the colors, the user could notice that the first sentence belonging to multiple topics, the second sentence shifted the topic to $sports$, and finally, from ``Meanwhile'' on, the topic is shifted to $techlology$ (and a little bit of $business$ given by the words ``investment'' or ``engage'' colored in green).
    Note that the user can also edit the document or even create new ones using the two buttons on the upper-right corner.}
    \label{fig:live_test}
\end{figure*}

\begin{figure*}[t!]
    \centering
    \includegraphics[width=190mm]{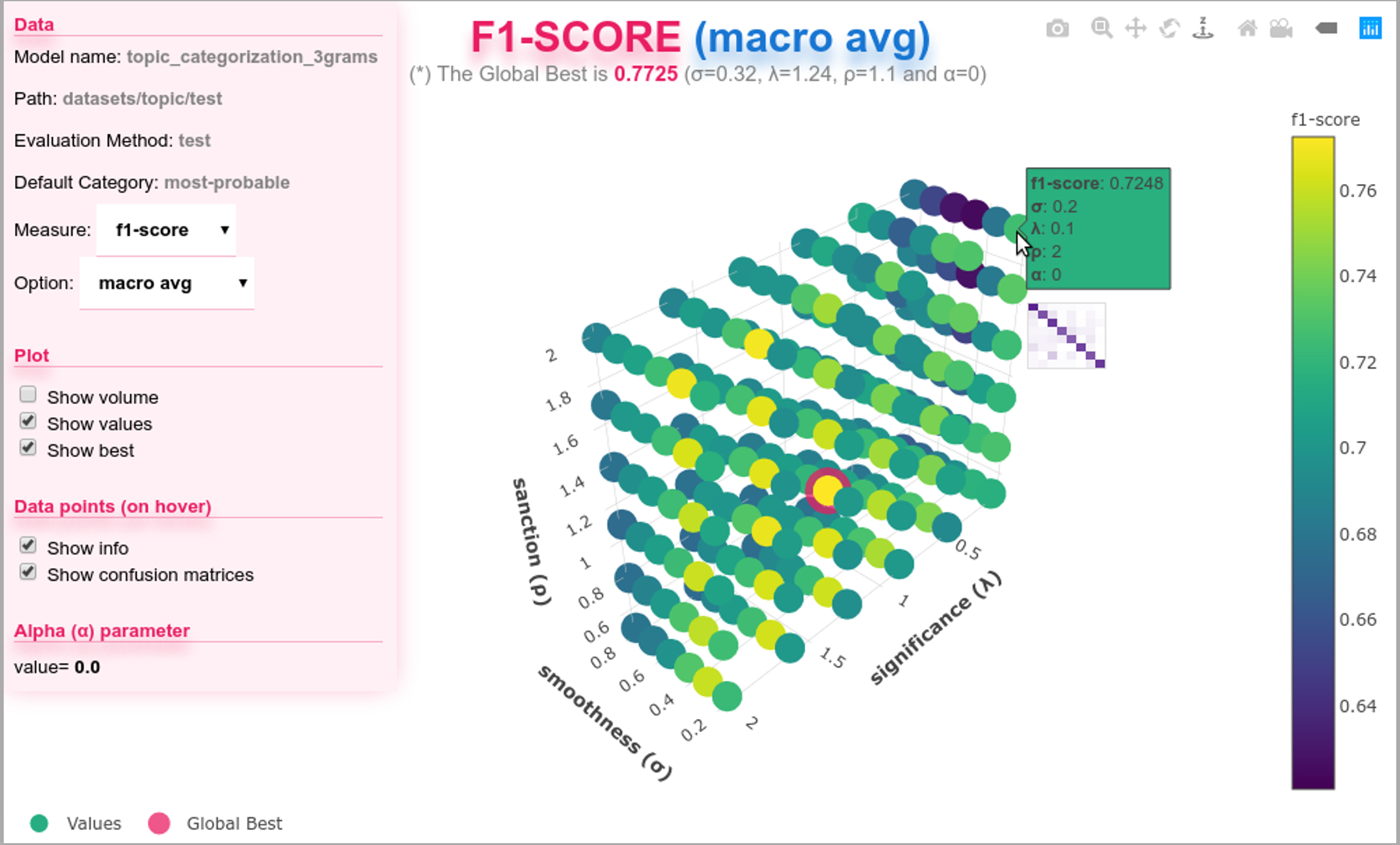}
    \caption{Evaluation plot screenshot. Each data point represents an evaluation/experiment performed using a particular combination of hyperparameter values. Points are colored proportionally to the obtained performance. The performance measure can be interactively changed using the options panel in the upper-left corner. Additionally, points with the global best performance are marked in pink. As shown in this figure, when the user moves the cursor over a point, information related to that evaluation is displayed, including a small version of the obtained confusion matrix.}
    \label{fig:eval_plot}
\end{figure*}

PySS3 is composed of one main module and three submodules. 
The main module is called ``pyss3'' and contains the classifier's implementation \emph{per se} in a class called ``SS3''.
The $SS3$ class implements not only the ``plain-vanilla'' version of the classifier \cite{burdisso2019} but also different variants, such as the one introduced later by the same authors \cite{burdisso2019-tss3}, which allows SS3 to recognize important word n-grams ``on the fly''.
Additionally, the $SS3$ class exposes a clear and simple API, similar to that of Scikit-learn models,\footnote{For instance, it has methods like ``fit'' for training and ``predict'' for classifying. Full list available in the API documentation: \href{https://pyss3.rtfd.io/en/latest/api/index.html\#pyss3.SS3}{https://pyss3.rtfd.io/en/latest/api/index.html\#pyss3.SS3}.} as the reader will notice in the example shown in \autoref{eg:1}. Finally, this module contains the following three submodules:

\begin{itemize}
    \item \emph{pyss3.server} --- contains the server's implementations for the ``Live Test'' tool, described in \autoref{subsec:soft-func}. An illustrative example of its use is shown in \autoref{eg:2}.
    \item \emph{pyss3.cmd\_line} --- implements the ``PySS3 Command Line'' tool, described in \autoref{subsec:soft-func}. This submodule is not intended to be imported and directly used with Python.
    \item \emph{pyss3.util} --- this submodule consists of a set of  utility and helper functions and classes, such as classes for loading data from datasets or preprocessing text.
\end{itemize}

\subsection{Implementation platforms}

PySS3 was developed using Python and was coded to be compatible with Python 2 and Python 3 as well as with different operating systems, such as Linux, macOS, and Microsoft Windows. To ensure this compatibility holds when the source code is updated, we have configured and linked the PySS3 Github repository with the Travis CI service. This service automatically runs the PySS3's test scripts using different operating systems and versions of Python whenever new code is pushed to the repository.\footnote{To monitor the compatibility state of the latest version of PySS3 online, visit \href{https://travis-ci.org/sergioburdisso/pyss3}{\url{https://travis-ci.org/sergioburdisso/pyss3}}}

\subsection{Software functionality}
\label{subsec:soft-func}

PySS3 is distributed via Python Package Index (PyPI) and therefore can be installed\footnote{For more details about installation, please refer to our on-line documentation: \href{https://pyss3.rtfd.io/en/latest/user_guide/installation.html}{\url{https://pyss3.rtfd.io/en/latest/user_guide/installation.html}}} simply by using the $pip$ command as follows:

\begin{minted}[
fontsize=\small,
bgcolor=console,
formatcom=\color{white}
]{bash}
$ pip install pyss3
\end{minted}

The package comes, in addition to the classifier, with two useful tools that allow working with SS3 in a very straightforward, interactive, and visual way, namely the ``Live Test'' tool and the ``PySS3 Command Line'' tool.

\

The ``Live Test'' tool is an interactive visualization tool that allows users to test his/her models ``on the fly''.
This tool can be launched with a single line of python code using the $Server$ class (see \autoref{eg:2} for an example).
The tool provides a user interface (a screenshot is shown in \autoref{fig:live_test}) by which the user can manually and actively test the model being developed using either the documents in the test set or just typing in his/her own. 
Also, the tool allows researchers to analyze and understand what their models are learning by providing an interactive and visual explanation of the classification process at three different levels (word n-grams, sentences, and paragraphs). 
We recommend trying out some of the online live demos available at \href{http://tworld.io/ss3}{\url{http://tworld.io/ss3}}.\footnote{Like, for instance, the demos for Topic Categorization or Sentiment Analysis on Movie Reviews.}

\

The ``PySS3 Command Line'' is an interactive command-line tool.
This tool allows users to deploy SS3 models and interact with them through special commands for every stage of the machine learning pipeline (such as model selection, training, or testing). Probably one of its most important features is the ability to automatically (and permanently) record the history of every evaluation that the user has performed (such as tests, k-fold cross-validations, or grid searches).\footnote{These features are also available through the $pyss3.util.Evaluation$ class.} This tool allows the user to visualize their models' performance in terms of the values of the hyperparameters, as shown in \autoref{eg:3}. This tool can be started from the operating-system command prompt using the ``pyss3'' command, automatically added to the system when installing the package.

\section{Illustrative examples}
\label{sec:examples}

\begin{figure}[t!]
    \centering
    \includegraphics[width=90mm]{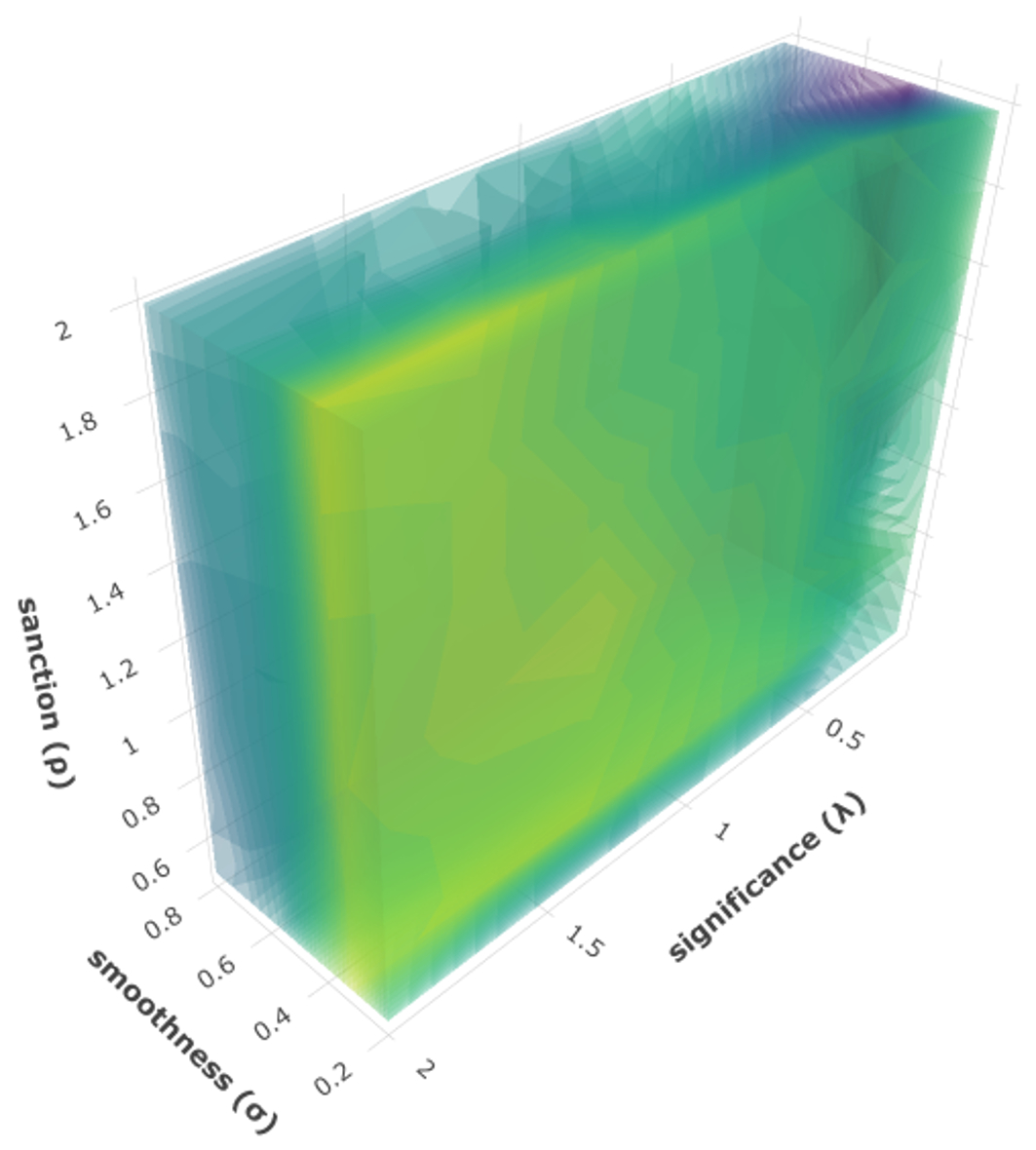}
    \caption{Evaluation plot - ``show volume'' option enabled.}
    \label{fig:eval_plot_volume}
\end{figure}

In this section, we will introduce three simple illustrative examples.
PySS3 provides two main types of workflow.
In the classic workflow, the user, as usual, imports the needed classes and functions from the package and then writes a python script to train and test the classifiers. In the ``command-line'' workflow, the whole machine learning pipeline is done using only the ``PySS3 Command Line'' tool, without coding in Python.
Due to space limitations, we will not show examples for the latter here. However, for full working examples using both workflows, please refer to the tutorials in the documentation (\href{https://pyss3.rtfd.io/en/latest/user_guide/getting-started.html#tutorials}{\url{https://pyss3.rtfd.io/en/latest/user_guide/getting-started.html\#tutorials}}).\footnote{Readers interested in trying PySS3 out right away, we have created Jupyter Notebooks for the tutorials, which can be used to interact with PySS3 in an online live environment  (\href{https://mybinder.org/v2/gh/sergioburdisso/pyss3/master?filepath=examples}{\url{https://mybinder.org/v2/gh/sergioburdisso/pyss3/master?filepath=examples}}).}

\

In the following examples, we will assume the user has already loaded the training and test documents and category labels, as usual, in the $x\_train$, $y\_train$, $x\_test$, $y\_test$ lists, respectively. For instance, this could be done using the $Dataset$ class from the $pyss3.util$ submodule, as follows:

\begin{minted}[
frame=lines,
framesep=2mm,
baselinestretch=1.2,
bgcolor=lightgray,
fontsize=\footnotesize,
% linenos
]{python}
from pyss3.util import Dataset

x_train,y_train = Dataset.load_from_files("path/to/train")
x_test, y_test = Dataset.load_from_files("path/to/test")
\end{minted}

\subsection{Training and test}
\label{eg:1}

This simple example shows how to train and test an SS3 model using default values.

\begin{minted}[
frame=lines,
framesep=2mm,
baselinestretch=1.2,
bgcolor=lightgray,
fontsize=\small,
% fontsize=\footnotesize,
% linenos
]{python}
from pyss3 import SS3

clf = SS3()
clf.fit(x_train, y_train)

y_pred = clf.predict(x_test)
print("Accuracy:", accuracy(y_pred, y_test))
\end{minted}

The last line prints the obtained accuracy, we are assuming here that this $accuracy$ function already exists.\footnote{For instance, it was previously imported from $sklearn.metrics$.}
Note that since SS3 creates a language model for each category, we do not need to create a document-term matrix, we are simply using the raw $x\_train$ and $x\_test$ documents for training and test, respectively.

\subsection{Training and (live) test}
\label{eg:2}

This example is similar to the previous one. However, instead of simply using $predict$ and $accuracy$ to measure our model's performance, here we are using the ``Live Test'' tool to analyze and test our model visually.

\begin{minted}[
frame=lines,
framesep=2mm,
baselinestretch=1.2,
bgcolor=lightgray,
fontsize=\small,
% linenos
]{python}
from pyss3 import SS3
from pyss3.server import Live_Test

clf = SS3()
clf.fit(x_train, y_train)

Live_Test.run(clf, x_test, y_test)
\end{minted}

\subsection{Hyperparameter optimization}
\label{eg:3}

This example shows how we could use the ``Evaluation'' class to find better hyperparameter values for the model trained in the previous example. Namely, we will perform hyperparameter optimization using the \emph{grid search} method, as follows:

\begin{minted}[
frame=lines,
framesep=2mm,
baselinestretch=1.2,
bgcolor=lightgray,
fontsize=\small,
% linenos
]{python}
from pyss3.util import Evaluation

best_s, best_l, best_p, _ = Evaluation.grid_search(
    clf, x_test, y_test,
    s=[0.2 , 0.32, 0.44, 0.56, 0.68, 0.8],
    l=[0.1 , 0.48, 0.86, 1.24, 1.62, 2],
    p=[0.5, 0.8, 1.1, 1.4, 1.7, 2]
)
\end{minted}

Note that in PySS3, the $\sigma$, $\lambda$, and $\rho$ hyperparameters are referenced using the ``s'', ``l'', and ``p'' letters, respectively. Thus, in this grid search, $\sigma$ will take six different values between .2 and .8, $\lambda$ between .1 and 2, and $\rho$ between .5 and 2. Once the grid search is over, the best hyperparameter values will be stored in those three variables. We can also use the ``plot'' function to visualize our results:

\begin{minted}[
frame=lines,
framesep=2mm,
baselinestretch=1.2,
bgcolor=lightgray,
fontsize=\small,
% linenos
]{python}
Evaluation.plot()
\end{minted}

This function will first save, in the current directory, a single and portable HTML file containing an interactive 3D plot. Then, it will open it up in the web browser; a screenshot is shown in \autoref{fig:eval_plot},\footnote{More info available in the documentation (\href{https://pyss3.rtfd.io/en/latest/user_guide/visualizations.html\#evaluation-plot}{\url{https://pyss3.rtfd.io/en/latest/user_guide/visualizations.html\#evaluation-plot}}).} 
in which we can see that the best hyperparameter values are $\sigma=0.32$, $\lambda=1.24$, and $\rho=1.1$. Finally, we will update the hyperparameter values of our already trained model using the ``set\_hyperparameters()'' function, as follows:

\begin{minted}[
frame=lines,
framesep=2mm,
baselinestretch=1.2,
bgcolor=lightgray,
fontsize=\small,
% linenos
]{python}
clf.set_hyperparameters(s=best_s, l=best_l, p=best_p)
\end{minted}

Alternatively, we could also create and train a new model using the obtained best values:

\begin{minted}[
frame=lines,
framesep=2mm,
baselinestretch=1.2,
bgcolor=lightgray,
fontsize=\small,
% linenos
]{python}
clf = SS3(s=0.32, l=1.24, p=1.1)
clf.fit(x_train, y_train)
\end{minted}

Note that, in addition to using the 3D evaluation plot to obtain the best values, users can use it to analyze (and better understand) the relationship between hyperparameters and performance in the particular problem being addressed. 
For instance, if the ``show volume'' option is enabled from the options panel, the plot will turn into the plot shown in \autoref{fig:eval_plot_volume}.
Using this plot, now, one can see that the \emph{sanction} ($\rho$) hyperparameter does not seem to impact performance.
In contrast, the performance seems to increase as the \emph{significance} ($\lambda$) value increases and seems to drop as the \emph{smoothness} ($\sigma$) hyperparameter moves away from $0.35$.
\footnote{It is worth mentioning that researchers could share these 3D model evaluation's portable files in their papers, which would increase experimentation transparency. For instance, we have uploaded the file used for the evaluation shown in \autoref{fig:eval_plot} for readers to interact with it: \href{https://pyss3.rtfd.io/en/latest/_static/eval_plot.html}{\url{https://pyss3.rtfd.io/en/latest/_static/eval_plot.html}}}

\section{Conclusions}
\label{sec:conclusions}

We have briefly presented PySS3, an open-source Python package that implements SS3 and comes with useful development and visualization tools.
This software could be useful for researchers and practitioners who need to deploy explainable and trusty machine learning models for text classification.



\appendix
\section{Required metadata}

\subsection{Current executable software version}

\autoref{tab:software} gives the information about the software release.

\begin{table}[!t]
\small
\begin{tabular}{l p{36mm} p{40mm}}
\hline
\textbf{Nr.} & \textbf{(executable) Software metadata description} &   \\
\hline
S1 & Current software version & v0.6.2 \\
S2 & Permanent link to executables of this version  &  \href{https://pypi.org/project/pyss3/0.6.2/}{\url{https://pypi.org/project/pyss3/0.6.2/}} \\
S3 & Legal Software License & MIT License \\
S4 & Computing platform/Operating System & Linux,
OS X,
and Microsoft Windows \\
S5 & Installation requirements \& dependencies & Pip,
Python 2.7-3.x,
Scikit-learn 0.20 or higher,
Matplotlib\\
S6 & Link to documentation &  \href{https://pyss3.readthedocs.io}{\url{https://pyss3.rtfd.io}} \\
S7 & Support email for questions & \href{mailto:sergio.burdisso@gmail.com}{\url{sergio.burdisso@gmail.com}} \\
\hline
\end{tabular}
\caption{Software metadata}
\label{tab:software} 
\end{table}

\subsection{Current code version}

\autoref{tab:code} describes the metadata about the source code of
PySS3.

\begin{table}[!t]
\small
\begin{tabular}{l p{36mm} p{40mm}}
\hline
\textbf{Nr.} & \textbf{Code metadata description} &  \\
\hline
C1 & Current code version & v0.6.2 \\
C2 & Permanent link to code/repository used of this code version & \href{https://github.com/sergioburdisso/pyss3}{\url{https://github.com/sergioburdisso/pyss3}} \\
C3 & Legal Code License   & MIT License \\
C4 & Code versioning system used & git \\
C5 & Software code languages, tools, and services used &  Python, Javascript, HTML\\
C6 & Compilation requirements, operating environments \& dependencies & Python 2.7-3.x,
Scikit-learn 0.20 or higher,
Matplotlib \\
C7 & Link to developer documentation/manual & \href{https://pyss3.readthedocs.io/en/latest/api/}{\url{https://pyss3.rtfd.io/en/latest/api}} \\
C8 & Support email for questions & \href{mailto:sergio.burdisso@gmail.com}{\url{sergio.burdisso@gmail.com}} \\
\hline
\end{tabular}
\caption{Code metadata}
\label{tab:code} 
\end{table}

\bibliographystyle{elsarticle-num}






\end{document}